\pdfoutput=1
\documentclass[final,3p,times,twocolumn]{elsarticle}

\usepackage{amssymb}
\usepackage{amsmath}
\usepackage{amsthm}
\usepackage{graphicx}
\usepackage{subcaption}
\usepackage{orcidlink}
\usepackage{float}
\usepackage{algorithm}
\usepackage{algpseudocode}
\usepackage{booktabs,siunitx}

\journal{Medical Image Analysis}
\makeatletter
\def\ps@pprintTitle{%
  \let\@oddhead\@empty \let\@evenhead\@empty
  \let\@oddfoot\@empty \let\@evenfoot\@empty}
\makeatother

\begin{document}

\begin{frontmatter}



\title{CPT-4DMR: Continuous sPatial-Temporal Representation for 4D-MRI Reconstruction}


\author[psi]{Xinyang Wu\fnref{eq}}
\author[psi,d-phys]{Muheng Li\fnref{eq}}
\author[psi,d-infk]{Xia Li\fnref{eq}}
\author[unibasp,uniba,ibt]{Orso Pusterla}
\author[psi]{Sairos Safai}
\author[uniba]{Philippe C. Cattin}
\author[psi,d-phys]{Antony J. Lomax}
\author[psi]{Ye Zhang\corref{cor}} 

\cortext[cor]{Corresponding author: ye.zhang@psi.ch}
\fntext[eq]{These authors contributed equally.} 

\affiliation[psi]{organization={Center for Proton Therapy, Paul Scherrer Institut},
            city={Villigen},
            country={Switzerland}}
\affiliation[d-phys]{organization={Department of Physics, ETH Zurich},
            city={Zurich},
            country={Switzerland}}
\affiliation[d-infk]{organization={Department of Computer Science, ETH Zurich},
            city={Zurich},
            country={Switzerland}}
\affiliation[unibasp]{organization={Department of Radiology, Division of Radiological Physics, University Hospital Basel},
            city={Basel},
            country={Switzerland}}
\affiliation[uniba]{organization={Department of Biomedical Engineering, University of Basel},
            city={Allschwil},
            country={Switzerland}}
\affiliation[ibt]{organization={Institute for Biomedical Engineering, University and ETH Zurich},
            city={Zurich},
            country={Switzerland}}

\begin{abstract}
Four-dimensional MRI (4D-MRI) is an promising technique for capturing respiratory-induced motion in radiation therapy planning and delivery. 
Conventional 4D reconstruction methods, which typically rely on phase binning or separate template scans, struggle to capture temporal variability, complicate workflows, and impose heavy computational loads. 
We introduce a neural representation framework that considers respiratory motion as a smooth, continuous deformation steered by a 1D surrogate signal, completely replacing the conventional discrete sorting approach. The new method fuses motion modeling with image reconstruction through two synergistic networks: the Spatial Anatomy Network (SAN) encodes a continuous 3D anatomical representation, while a Temporal Motion Network (TMN), guided by Transformer-derived respiratory signals, produces temporally consistent deformation fields.
Evaluation using a free-breathing dataset of 19 volunteers demonstrates that our template- and phase-free method accurately captures both regular and irregular respiratory patterns, while preserving vessel and bronchial continuity with high anatomical fidelity.
The proposed method significantly improves efficiency, reducing the total processing time from approximately five hours required by conventional discrete sorting methods to just 15 minutes of training. Furthermore, it enables inference of each 3D volume in under one second. 
The framework accurately reconstructs 3D images at any respiratory state, achieves superior performance compared to conventional methods, and demonstrates strong potential for application in 4D radiation therapy planning and real-time adaptive treatment.

\end{abstract}




\end{frontmatter}



\section{Introduction}
Four-dimensional magnetic resonance imaging (4D-MRI)  has emerged as an promising technique to capture respiratory-induced anatomical changes for motion management in radiotherapy \cite{paganelli2018mri}. Unlike static 3D imaging, 4D-MRI produces a dynamic organ-motion representation with superior soft-tissue contrast compared to 4D-CT, which is especially beneficial for tracking abdominal tumors in real-time. This modality also offers flexible imaging contrasts (T1-weighted \cite{deng2016four}, T2-weighted \cite{freedman2018super}, hybrid T2/T1 \cite{liu2016accuracy}) and customizable scan parameters \cite{khoo2006new}. Thanks to MRI’s non-ionizing nature, clinicians can acquire longer image sequences covering many breathing cycles, yielding a representative “average” motion pattern \cite{national2006health}. This capability also enables a more precise localization of moving targets during the radiation dose delivery.

Despite many advantages of 4D-MRI, three major challenges continue to impede its clinical applications. (1) Conventional reconstruction approaches are heavily based on respiratory phase binning, typically assuming periodic breathing patterns and reconstructing only a single respiratory cycle \cite{cai2011four}. This phase-dependent methodology cannot adequately account for the inherent variability between multiple breathing cycles, making it challenging to accurately represent long-term respiratory motion in realistic clinical scenarios, particularly during prolonged free-breathing acquisitions. 
(2) Current deep learning approaches for 4D-MRI reconstruction add complexity by requiring a pre-acquired template volume before learning motion patterns \cite{qiu2024stnerpspatialtemporalneuralrepresentation}. This need for a static reference scan not only complicates the imaging protocol (e.g., requiring an extra breath hold or gating sequence), but also demands more complex preprocessing and training strategies.
(3) Conventional sorting-based 4D reconstruction is computationally intensive and time-consuming \cite{von20074d}. These methods require performing many non-rigid registrations (on the order of hours of processing per patient), which makes real-time or even same-day 4D-MRI reconstruction impractical for future online adaptive therapy.
These factors raise the barrier to clinical implementation.

To overcome these limitations, we propose a template-free continuous representation framework for 4D-MRI reconstruction that significantly simplifies data acquisition and reconstruction. 
(1) To overcome the limitations of phase binning, our method leverages a continuous neural representation for the anatomy and motion modeling. 
Guided by a phase-free respiratory surrogate signal (e.g., derived from diaphragm motion), it can reconstruct arbitrary respiratory states across multiple, variable breathing cycles.  
(2) By jointly optimizing the anatomy and motion on the fly, our approach eliminates the need for a pre-acquired template volume, which simplifies the imaging protocol. 
(3) To address the prohibitive reconstruction time of conventional methods, our framework takes advantage of the compactness and training speed of an implicit neural representation (INR). This reduces the entire reconstruction time to under one second, a significant reduction from several hours required previously, which opens the potential for advanced image guidance and real-time adaptive radiotherapy. 


At the core of our proposed framework is the joint optimization of two coupled implicit neural networks: one learns a continuous 3D anatomy (a neural representation of the volume), and the other learns a time-resolved continuous deformation field that maps any respiratory state to a reference state. This integrated model transforms sparsely acquired data into a complete and continuous spatio-temporal representation.
We can thus sample high-fidelity volumes at arbitrary respiratory positions or time stamps, effectively capturing both regular and irregular breathing patterns with high accuracy and efficiency. Sparse, discontinuous slice acquisitions are transformed into a complete, temporally coherent volumetric representation of the patient that can be queried in real-time for any respiratory state. Our contributions include:
\begin{enumerate}
    \item[$\bullet$] We acquired an extensive MRI dataset for 4D reconstruction, consisting of 19 volunteers, each with a 10-minute free-breathing acquisition.
    \item[$\bullet$] We introduced a template-free, phase-free neural representation framework for 4D-MRI that jointly optimizes anatomical representation and temporal deformation fields for consistent reconstruction across multiple breathing cycles.
    \item[$\bullet$] We designed a novel Temporal Motion Network that learns continuous deformation vector fields (DVFs) from spatial coordinates and a respiratory surrogate signal, effectively handling irregular breathing patterns and inter-cycle variations without requiring a prior motion model or phase assumptions.
    \item[$\bullet$] We demonstrated the clinical feasibility of our method through extensive validation on a dataset of 19 volunteers, achieving arbitrary-phase reconstruction in real-time with temporal coherence, high anatomical fidelity, and spatial continuity.
\end{enumerate}
\section{Related Works}
\label{related_work}
\subsection{Conventional 4D-MRI Reconstruction}
Conventional 4D-MRI separates image formation into acquisition and reconstruction, each shaped by respiratory motion and data-throughput constraints. Two main acquisition paradigms exist. Time-resolved multi-slice 2D imaging collects sequential (often interleaved) slices that are later resorted into respiratory bins, by phase or amplitude, using external bellows, optical markers, or internal navigator/self-gating metrics \cite{perkins2021experimental,van2018self}. Axial stacks deliver high in-plane detail but suffer inter-plane artifacts, so sagittal or coronal prescriptions are frequently preferred \cite{bernstein2004handbook}. Volumetric 3D imaging, in contrast, excites the entire field of view with every RF pulse, encoding all three spatial dimensions in k-space to yield isotropic voxels, contiguous volumes, and flexible trajectories; however, its short-TR demands confine contrast largely to T1-weighted gradient-echo or balanced SSFP sequences \cite{thomas2022contrast}. Because fully sampled 4D data would be prohibitively large, most protocols undersample k-space, forcing the reconstruction stage to compensate. The prevailing reconstruction pipeline, slice stacking, registers 2D slices from different breathing instants, fits non-rigid B-spline deformations to warp each slice into a common respiratory frame, and then concatenates the motion-corrected slices to build a 3D volume for each bin \cite{von20074d}. Repeating this across bins yields the four-dimensional series, but the approach falters when breathing is irregular and is computationally onerous, typically requiring about five hours of registration and sorting per patient. These trade-offs between data completeness, artefact suppression, contrast flexibility, and computational burden define the current landscape and limitations of clinical 4D-MRI.

\subsection{Deep Learning-Augmented 4D Reconstruction}
Recent advances in computer vision, most notably Neural Radiance Fields (NeRF) \cite{nerf2020} and 3D Gaussian Splatting (3DGS) \cite{kerbl2023gaussian}, have shown that high-fidelity four-dimensional scenes can be reconstructed by coupling temporal dynamics with spatial representations. When adapted to medical imaging, deep-learning reconstructions routinely outperform conventional pipelines in artefact suppression and detail preservation \cite{hammernik2021systematic, knoll2020deep}; however, clinical deployment is still constrained by heavy computation, the difficulty of modeling respiratory deformations, and the undersampling that characterizes most k-space acquisitions \cite{gao2024nerfmedical}. Modality-specific solutions have begun to mitigate these issues. DraculaNet accelerates 4DCT reconstruction with a deep radial 3D U-Net while preserving tumour delineation accuracy \cite{chen2023dracula}; US-4D-MRI reduces MRI scan times by steering reconstruction with ultrasound-derived diaphragm motion surrogates \cite{arnold2019us4dmri}; and FlowVN incorporates physics-based priors into a variational network to recover four-dimensional flow fields rapidly \cite{liu2021flowvn}. Despite these advances, most methods still rely on template-based motion models or bespoke acquisition protocols, which limits generalisability and keeps real-time interventional use, where sub-second visualisation of respiratory motion is critical, out of reach \cite{wang2022mahdtv,johnson2022flowmrinet,hu2023transfer}.

\subsection{Implicit Neural Representation for Image Reconstruction}
Implicit neural representations (INRs) have emerged as powerful tools in medical imaging for parameterizing signals as continuous functions~\cite{molaei2023implicit}. Research in this area has evolved along two main lines: deformable image registration and tomographic reconstruction. The first focuses on modeling motion within the image domain, where INRs represent continuous deformation fields~\cite{wolterink2022implicit,van2023robust}. This has progressed to modeling spatio-temporal velocity fields, enabling the capture of complex biological motion~\cite{li2024cptdir}. In parallel, a second line of research has focused on tomographic reconstruction from sparse data, but operates almost exclusively in the projection domain. These methods use a differentiable forward model (e.g., Radon transform) to compare the INR output with raw sensor data~\cite{shi2024implicit, lee2024iterative, zhang2023dynamic, reed2021dynamic}. However, this approach is unsuitable for image-domain data and often relies on prior images to constrain the ill-posed problem~\cite{shen2022nerp, song2023piner}. A critical gap therefore exists, as no method performs 4D reconstruction directly from sparse, image-domain slices in a template-free manner. Our work, CPT-4DMR, bridges this gap by fusing the strengths of both research lines to achieve template-free 4D-MRI reconstruction solely from 2D slice data.



\section{Methodology}

\begin{figure*}[htbp]
    \centering
    \includegraphics[width=\textwidth]{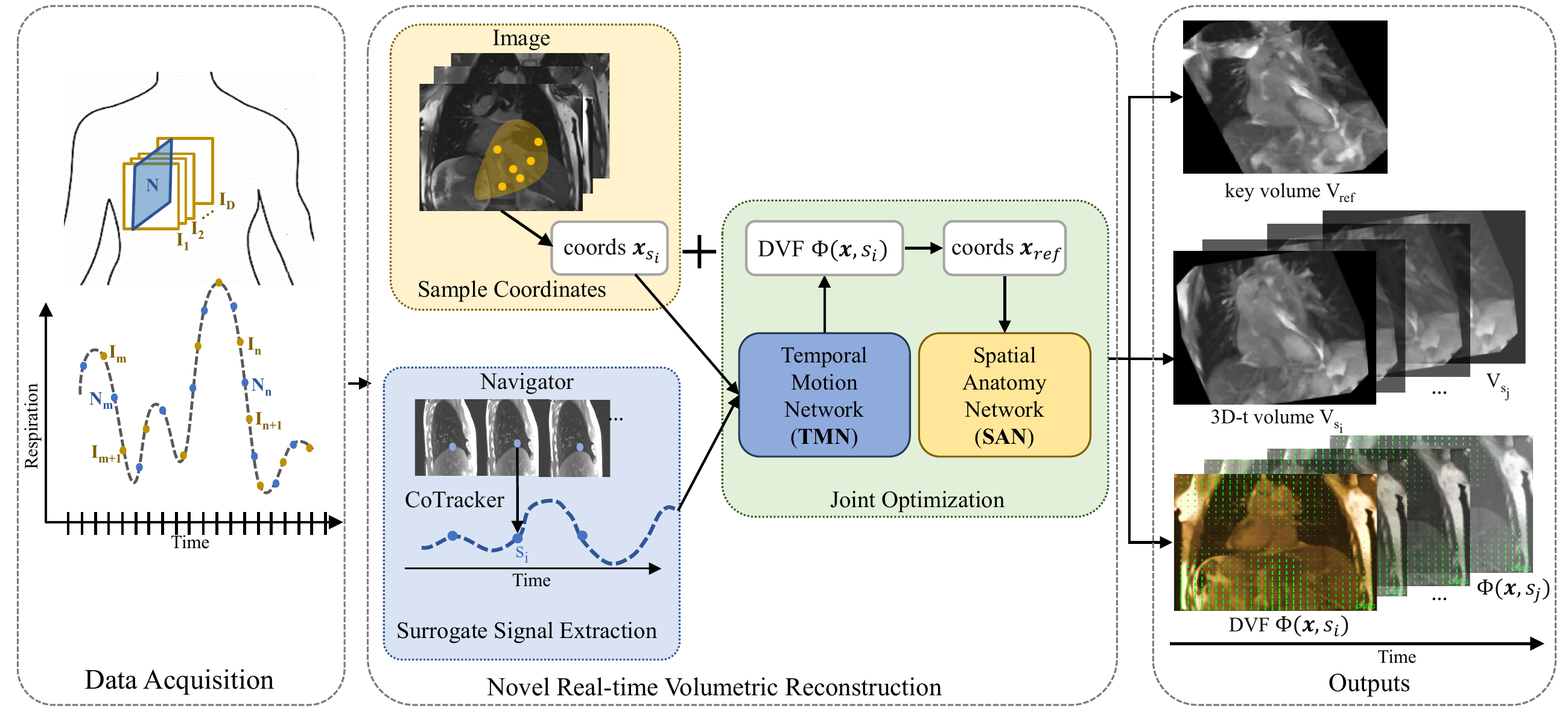}
    \caption{The overview of our method. We randomly sample around 10,000 coordinate points each breathing state from image slices, Temporal Motion Network will calculate the DVF wrapping from current state $s_i$ back to template state $ref$ based on the surrogate signal, add DVF to our coordinate $\textbf{x}_{s_i}$, we can get coordinates of the template $\textbf{x}_{ref}$, where we learns the relationship between each continuous coordinate and intensity $I_{ref}$.}
    \label{fig:method}
\end{figure*}

The core of our CPT-4DMR framework is the deconstruction of the 4D spatio-temporal problem. We separately model the patient's spatial anatomy and temporal motion using two dedicated neural networks, and then jointly optimize these representations to form a single, continuous function for 4D-MRI reconstruction. As illustrated in Figure \ref{fig:method}, these two synergistic components are: (i) A Spatial Anatomy Network learns a continuous representation of the canonical anatomy, effectively creating a high-fidelity, internal reference volume from the data itself.
(ii) A Temporal Motion Network models the respiratory dynamics by learning a mapping from a spatio-temporal location $(x,y,z,t)$ to a time-varying deformation vector field (DVF). This DVF warps the learned reference anatomy to any desired respiratory state at time $t$.
The entire model is trained end-to-end, requiring only the sparsely acquired 2D MRI slices and the associated surrogate signal. This obviates the need for a pre-acquired template volume, a distinguishing feature from methods like that of Qiu et al. \cite{qiu2024stnerpspatialtemporalneuralrepresentation}, which simplifies the clinical workflow. It is important to emphasize that our framework is entirely patient-specific: both the anatomical and motion models are trained individually for each patient without using any prior data from other patients. This patient-specific approach avoids potential hallucinations and prevents the networks from incorporating anatomical features from other patients when filling data gaps.


\subsection{Data Acquisition and Preprocessing}
\subsubsection{MRI Dataset and Imaging Protocol}
We acquired a comprehensive 4D-MRI dataset from 19 healthy volunteers, each undergoing three separate 10-minute free-breathing scans. The study was approved by the Ethikkommission Nordwest- und Zentralschweiz (Switzerland) under approval number EKNZ2019-01060. Data were collected on a 1.5T MAGNETOM Aera scanner (Siemens Healthineers, Erlangen, Germany) using a custom-developed balanced steady-state free precession (bSSFP) sequence with Cartesian k-space sampling~\cite{bauman2016ultra, bieri2013ultra}.
The dynamic imaging was performed with an interleaved cine protocol designed to capture respiratory motion while simultaneously generating a validation signal. This protocol alternated between acquiring a moving coronal slice stack through the lungs and a fixed sagittal navigator slice at a constant anatomical position~\cite{von20074d, peteani2024retrospective}. This orthogonal plane setup ensured that the navigator data provided an independent measure of motion, while the coronal series offered a comprehensive view of thoracic dynamics.
The coronal acquisition produced volumes of 384 × (272–328) × 64 voxels (most commonly 384 × 300) with an in-plane resolution ranging from 1.04 × 1.04 mm² to 1.28 × 1.28 mm² (mean: 1.092 mm). The navigator images had a similar in-plane resolution. Both sequences used a slice thickness of 7 mm and a flip angle between 45° and 53° (mean: 50.2°). The acquisition employed identical echo times (TE = 1.09 ms) across all scans, while acquisition times were either 313.59 ms or 330.94 ms, depending on the scan setup.  A complete list of acquisition parameters is provided in Table~\ref{tab:acquisition_params}. Based on acquisition parameters (e.g., number of phase encodes, partial Fourier 7/8, GRAPPA factor 2 with 24 reference lines), the effective bSSFP sequence TR corresponds to approximately 2.6–2.9 ms per segment, and the listed TR value reflects the segment acquisition time.

\begin{table}
\caption{Summary of MRI acquisition parameters.}
\label{tab:acquisition_params}
\centering
\begin{tabular}{l|c}
\toprule
\textbf{Parameter} & \textbf{Value} \\
\hline
Pixel Spacing (mm) & 1.04-1.28 (mean: 1.092) \\
Slice Spacing (mm) & 3.29-4.20 (mean: 3.728) \\
Slice Thickness (mm) & 7 \\
TE (ms) & 1.09 \\
Segment TR (ms) & 2.6-2.9 \\
Acquisition Time (ms) & 313.59-330.94\\
Flip Angle (degrees) & 45-53 (mean: 50.2) \\
\bottomrule
\end{tabular}
\end{table}

\subsubsection{Respiratory Surrogate Signal Extraction}

Our Temporal Motion Network requires a respiratory surrogate signal as input to establish the temporal correspondence between different breathing states during reconstruction. Rather than relying on external monitoring devices (such as respiratory belts or optical markers), we derive an internal surrogate signal $s(t)$ directly from the acquired sagittal navigator slices to represent the patient's respiratory state over time. This approach eliminates the need for additional hardware while providing a robust respiratory signal derived from the imaging data itself. Instead of performing time-consuming non-rigid registration, we track diaphragm landmarks with CoTracker \cite{karaev2024cotracker}, a Transformer \cite{vaswani2017attention} based algorithm that accurately follows thousands of 2D features in long image sequences. We select five diaphragm landmarks, average their vertical displacements, and normalize the result to $[0,1]$ to obtain the surrogate signal $s(t)$. This internal surrogate strategy cuts slice-sorting time from about five hours to five minutes and provides the Temporal Motion Network with a normalized respiratory state $s$ that is concatenated with spatial coordinates $(x,y,z)$ as input for deformation field prediction.



\subsection{Continuous sPatial and Temporal Modeling}
Our model parameterizes the 4D MRI sequence as a function that maps a spatial coordinate $\mathbf{x}=(x,y,z)$ and a respiratory state $s$ to a corresponding voxel intensity $I$. This is achieved through two specialized neural networks built upon the Sinusoidal Representation Network (SIREN) architecture \cite{sitzmann2020implicit}, which uses periodic sine functions as activations and is well-suited for representing complex signals and their derivatives.

\subsubsection{Temporal Motion Network}

The Temporal Motion Network (TMN) learns a continuous mapping from a given respiratory state $s$ to a 3D DVF $\Phi$. Its function is to predict the coordinate shifts that align any given respiratory state with a canonical reference state $s_{\text{ref}}$. This process is formulated as a regression problem, where the network $f_{\text{TMN}}$ takes a spatial coordinate vector $\mathbf{x} = (x,y,z)$ and a scalar state $s$ as input to predict the corresponding displacement vector:

\begin{equation}
    \Phi(\mathbf{x}, s) = f_{\text{TMN}}(\mathbf{x}, s)
\end{equation}

where $\Phi$ is the DVF. This allows us to map a point $\mathbf{x}$ from an observed state $s$ back to its corresponding location $\mathbf{x}_{\text{ref}}$ in the reference frame:
\begin{equation}
    \mathbf{x}_{\text{ref}} \leftarrow \mathbf{x} + \Phi(\mathbf{x}, s) \label{eq:warp}
\end{equation}

This design effectively models complex and irregular respiratory motion, a known challenge for conventional 4D-MRI methods \cite{ge2013planning}.
The network's architecture is a specialized SIREN \cite{sitzmann2020implicit} variant, which we term \textbf{DeformShrink}. It is implemented as a 4-layer multilayer perceptron with 256 hidden units per layer. The periodic nature of SIREN's sinusoidal activations makes it highly effective at modeling the complex, high-frequency details of respiratory-induced deformations \cite{wolterink2022implicit}.

The input to the network, $\textbf{X}_{\text{+}} \in \mathbb{R}^{B \times N \times 4}$, is formed by concatenating the batch of 3D spatial coordinates ($\textbf{X} \in \mathbb{R}^{B \times N \times 3}$) with the corresponding normalized respiratory surrogate state, $\tilde{s}$. Note that the raw surrogate signal, derived from the mean z-axis displacement of tracked diaphragm landmarks, is specifically normalized to the range $[-1, 1]$ for network input.

A key feature of our DeformShrink architecture is the application of the $\text{Tanhshrink}$ activation function to the output of the SIREN backbone. The complete operation is:
\begin{equation}
    f_{\text{TMN}}(\textbf{x}, s) = \text{Tanhshrink}(f_{\text{SIREN}}(\textbf{x}, \tilde{s}))
\end{equation}
where $f_{\text{SIREN}}$ represents the main SIREN network. The $\text{Tanhshrink}$ function, defined as:
\begin{equation}
    \text{Tanhshrink}(\textbf{z}) = \textbf{z} - \tanh(\textbf{z}) \label{eq:tanhshrink_new}
\end{equation}
is crucial for keeping the learned motion robust. It effectively suppresses small, potentially noisy deformations in static areas and removes unnecessary spurious oscillatory movements, while preserving significant, physically plausible motion patterns.

\subsubsection{Spatial Anatomy Network}



The Spatial Anatomy Network (SAN) provides a high-fidelity representation of the patient's base anatomy. It functions as a continuous implicit neural representation (INR), learning a mapping $f_{\text{SAN}}$ from a 3D coordinate $\mathbf{x}_{\text{ref}}$ in the canonical reference frame to its corresponding MR signal intensity $I$:
\begin{equation}
    I_{\text{pred}}(\mathbf{x}_{\text{ref}}) = f_{\text{SAN}}(f_{\text{SIREN}}(\textbf{x}_{\text{ref}}) + \text{Skip}(\textbf{x}_{\text{ref}})),
\end{equation}
During reconstruction, coordinates from any state $s_i$ are first warped to the reference frame using the Temporal Motion Network, and then queried by the SAN to yield the final intensity.

To capture fine anatomical details while ensuring stable training, the network, which we term \textbf{SirenRes}, is implemented as a 5-layer SIREN with 512 hidden units per layer. Its architecture incorporates two key modifications to the standard SIREN backbone. First, a residual connection is introduced at the 4th layer to improve gradient flow through the deep network, which mitigates the vanishing gradient problem and enhances feature propagation. Second, the final layer uses a linear transformation followed by a sigmoid activation function, $\sigma(\cdot)$, to map the output to the normalized intensity range $[0, 1]$.

The layer-wise operation of SirenRes, where $\mathbf{h}_l$ is the output of layer $l$, can be expressed as:
\begin{equation}
\mathbf{h}_l =
\begin{cases}
    \sin(\omega_l \mathbf{W}_l \mathbf{h}_{l-1} + \mathbf{b}_l) & \text{for } l \in \{1, 2, 3\} \\
    \sin(\omega_4 \mathbf{W}_4 \mathbf{h}_{3} + \mathbf{b}_4) + \mathbf{h}_3 & \text{for } l = 4 \\
    \sigma(\mathbf{W}_5 \mathbf{h}_{4} + \mathbf{b}_5) & \text{for } l = 5
\end{cases}
\end{equation}
where $\mathbf{h}_0 = \mathbf{x}_{\text{ref}}$ is the input coordinate. This architecture allows for the effective learning of a detailed and continuous anatomical representation.

\subsection{Training and Reconstruction}
\subsubsection{Joint Optimization and Loss Function}

The Spatial Anatomy Network $f_{\text{TMN}}$ and the Temporal Motion Network $f_{\text{SAN}}$ are optimized jointly in an end-to-end manner. Unlike methods requiring a pre-acquired template \cite{qiu2024stnerpspatialtemporalneuralrepresentation}, our framework learns the canonical anatomy and the temporal deformations simultaneously from scratch, using only the acquired 2D slices and their corresponding surrogate signals. During training, we randomly sample batches of approximately 10,000 3D coordinate-intensity pairs $(\mathbf{x}, I(\mathbf{x}))$ from the ground-truth image slices for each optimization step.

The joint optimization is guided by a composite loss function, $\mathcal{L}_{\text{total}}$, which combines a photometric data-fidelity term with a smoothness regularization term for the motion field. The total loss is defined as:
\begin{equation}
    \mathcal{L}_{\text{total}} = \mathcal{L}_{\text{photo}} + \lambda \cdot \mathcal{L}_{\text{smooth}}
\end{equation}
where $\lambda$ is a hyperparameter balancing the two terms.

The photometric loss, $\mathcal{L}_{\text{photo}}$, ensures that the predicted intensity $I_{\text{pred}}$ matches the ground-truth intensity $I_{gt}$. The predicted intensity is obtained by first warping the sample coordinate $\mathbf{x}$ with the DVF from the Temporal Motion Network and then querying the resulting coordinate in the Spatial Anatomy Network. We use the L1 loss for this term:
\begin{equation}
    \mathcal{L}_{\text{photo}} = \frac{1}{N} \sum_{i=1}^{N} \left| f_{\text{SAN}}\left(\mathbf{x}_i + f_{\text{TMN}}(\mathbf{x}_i, s_i)\right) - I_{gt}(\mathbf{x}_i, s_i) \right|
\end{equation}
where $N$ is the number of sampled points in the mini-batch.

\begin{figure*}
   \centering
   \includegraphics[width=\textwidth]{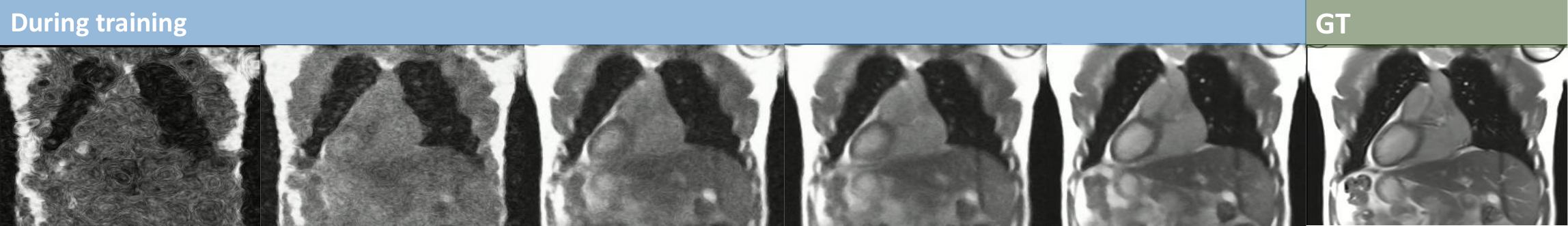}
   \caption{Progressive refinement of the canonical anatomical template during training, showing a coronal slice from volunteer 18 reconstructed at multiple epochs. This demonstrates that the network gradually converges to a high-fidelity reference volume through joint optimization.}
   \label{fig:training_progress}
\end{figure*}

The smoothness regularizer is designed to ensure the transformation is physically plausible. It penalizes local volume changes and prevents topological inconsistencies (i.e., tissue folding) by constraining the Jacobian determinant of the transformation map $\Phi$. The loss is formulated to encourage the determinant to be close to 1 (volume preservation) and implemented as in IDIR \cite{wolterink2022implicit}:
 
\begin{equation}
    \mathcal{L}_{\text{jacdet}} = \frac{1}{N} \sum_{i=1}^N|1-\det\nabla\Phi_i|
\end{equation}
This regularization is critical for generating coherent motion fields. Figure \ref{fig:training_progress} shows representative 2D slices reconstructed at different epochs. As training progresses, the anatomical details become increasingly well-defined, confirming that the network learns a stable, high-fidelity reference volume in a self-supervised manner, without requiring a static pre-acquired template.

\subsubsection{Real-Time Volumetric Reconstruction}
Once the networks are trained, a full 3D volume for any arbitrary respiratory state $s'$ can be reconstructed in real-time (defined here as sub-second reconstruction time suitable for adaptive radiotherapy workflows). The procedure leverages the learned spatio-temporal decomposition by first using the Temporal Motion Network $f_{\text{TMN}}$ to predict a deformation vector field (DVF) that maps each coordinate from the target state $s'$ to its corresponding location in the canonical reference frame. The Spatial Anatomy Network $f_{\text{SAN}}$ is then queried at these deformed coordinates to yield the final intensity values. The process is formally described as:
\begin{equation} 
    I(\textbf{x}, s') = f_{\text{SAN}}(\textbf{x} + f_{\text{TMN}}(\textbf{x}, s))
\end{equation}
For a practical implementation, the entire volume is reconstructed by processing a uniform grid of coordinates in batches. A key step involves normalizing the predicted DVF, which is initially in voxel units, to the range of $[-1,1]$. This normalization is required to align with the input conventions of standard grid-sampling functions in deep learning frameworks. The complete procedure is detailed in Algorithm 1.

\begin{algorithm}[H]
\caption{Novel Respiratory State Reconstruction}
\label{alg:reconstruction}
\begin{algorithmic}[1]
\Require Respiratory state $s'$, trained networks $f_{\text{TMN}}$ and $f_{\text{SAN}}$
\Ensure Reconstructed volume $\textbf{I}'$
\State $\text{pred\_slices} \gets []$
\For{each batch of coordinates $\mathbf{x}_{\text{grid}}$ in volume grid}
    \State $\Phi \gets f_{\text{TMN}}(\mathbf{x}_{\text{grid}}, s')$ \Comment{Predict DVF for grid coordinates}
    \State $\tilde{\Phi}_{\text{norm}} \leftarrow \text{normalize}(\Phi)$ \Comment{Scale DVF to coordinate range $[-1, 1]$}
    \State $\mathbf{x}_{\text{warped}} \leftarrow \mathbf{x}_{\text{grid}} + \tilde{\Phi}$ \Comment{Compute warped coordinates}
    \State $I' \leftarrow f_{\text{SAN}}(\mathbf{x}_{\text{warped}})$ \Comment{Query anatomy at warped locations}
    \State $\text{pred\_slices.append}(I')$
\EndFor
\State $\textbf{I}' \gets \text{assemble\_from\_points}(\text{pred\_slices})$ \Comment{Map points back to volume}
\State \Return $\textbf{I}'$
\end{algorithmic}
\end{algorithm}

The final assembly step maps the predicted point-wise intensities back to their original grid positions. The continuous nature of our representation ensures smooth transitions between respiratory states, maintaining anatomical plausibility throughout the reconstructed 4D sequence. This allows for the reconstruction of novel respiratory phases not present in the training data, including interpolated and extrapolated states. This capability is particularly valuable for applications requiring high temporal resolution, such as motion-adaptive radiation therapy planning and delivery.




\section{Experiments}

\subsection{Implementation Details}
We trained both networks for 10,000 epochs, which took roughly 15 min on a single NVIDIA RTX 4090 GPU with a batch size of 8. Throughout training, the Spatial Anatomy Network learns a continuous mapping from template-space coordinates to voxel intensities, producing a high-fidelity anatomical model with clearly resolved vessels (see Figure \ref{fig:vessel_continuity}). In parallel, the Temporal Motion Network learns to warp any respiratory state to the template, so a three-dimensional volume $V_{s_i}$ can be reconstructed in real time for any state $s_i$. 
The Motion Detector (MD) contains four hidden layers with 256 units each and sinusoidal activations, using $\omega_0 = 30$ in the first layer and $\omega_1 = 1$ in subsequent layers. The Implicit Representation Network (IRN) employs a deeper SirenRes backbone with five hidden layers of 512 units. 

Training was performed using the Adam optimizer with a learning rate of 5e-5, $\beta_1 = 0.9$, $\beta_2 = 0.999$. The batch size of 8 refers to processing 8 coordinate sampling batches simultaneously, where each batch contains 10,000 3D coordinate points sampled from the slice data. The networks are implemented as Multi-Layer Perceptrons (MLPs) with sinusoidal activations as described in Section 3.2.

\subsection{Evaluation Methods}
\label{sec:evaluation_methods}

We comprehensively evaluated our proposed framework through multiple strategies assessing clinical applicability and technical performance across reconstruction efficiency, motion accuracy, and volumetric quality.

\subsubsection{Data Partitioning and Sampling Strategy}

For each volunteer, we employed temporal data splitting using the first 11/12 of each sequence (704 slices) for training and the remaining 1/12 (64 slices) for validation. While evaluating on entire slice positions rather than temporal subsets could provide additional insights into spatial generalization, our current approach ensures evaluation on unseen respiratory states while maintaining breathing pattern continuity. This approach ensures evaluation on unseen respiratory states while maintaining breathing pattern continuity. During training and evaluation, we randomly sample approximately 10,000 3D coordinate points from each slice to balance computational efficiency with reconstruction fidelity.

\subsubsection{Evaluation Framework}

\textbf{Reconstruction Efficiency}: We measured inference time per 3D volume reconstruction ($384 \times 300 \times 64$) after training completion, comparing against traditional sorting-based methods~\cite{peteani2024retrospective}. The reported speed improvement primarily stems from eliminating the need for extensive non-rigid registration procedures, though the point tracking component also contributes to overall efficiency gains.

\textbf{Motion Accuracy}: 
Ground truth motion trajectories were extracted from navigator slices using CoTracker to track 5 anatomical landmarks on the diaphragm. Corresponding trajectories were extracted from reconstructed navigator slices using the same tracking approach and compared via Mean Absolute Error (MAE) computation between trajectory pairs, with motion amplitudes normalized to $[0,1]$ range.

\textbf{Volumetric Quality}: 
Volumetric reconstruction quality was assessed through direct pixel-wise comparison between ground truth and reconstructed image slices at identical spatial-temporal locations. Anatomical continuity was assessed through Maximum Intensity Projections (MIP) and temporal consistency evaluation across consecutive respiratory states.


\subsubsection{Quantitative Metrics}

Motion tracking was evaluated using ground truth diaphragm motion trajectories extracted from navigator slices as reference data. Quantitative assessment employed the Mean Absolute Error (MAE) between ground truth and reconstructed trajectory pairs (after re-normalizing both trajectories to the $[0,1]$ range for evaluation, distinct from the $[-1,1]$ normalization used during network training) and the Structural Similarity Index Measure (SSIM) for pattern comparison.
Image reconstruction quality was assessed using MAE, Mean Squared Error (MSE), and Peak Signal-to-Noise Ratio (PSNR), with intensities normalized to the $[0,1]$ range to ensure dimensionless, dataset-comparable values. Temporal coherence was measured through frame-to-frame consistency and deformation field regularity.
Motion accuracy evaluation focused primarily on five diaphragm landmarks due to their reliable trackability in navigator slices. Although incorporating additional liver landmarks would be desirable, this was limited by the inability to guarantee consistent and complete liver anatomy coverage across the imaging volume.

\begin{table}
   \caption{Quantitative evaluation of our INR-based 4D-MRI reconstruction compared to traditional sorting-based methods.}
   \label{tab1}
   \centering
   \resizebox{\columnwidth}{!}{%
   \begin{tabular}{l | c c c}
   \toprule
   \textbf{Method} & \textbf{MAE (navi)} $\downarrow$ & \textbf{MAE (img)} $\downarrow$ & \textbf{MSE (img)} $\downarrow$ \\
   \hline
   Traditional Sorting & 0.24±0.06 & 0.10±0.02 & 0.03±0.01 \\
   Our Method          & 0.12±0.06 & 0.05±0.01 & 0.01±0.00 \\
   \bottomrule
   \end{tabular}%
   }
\end{table}



\subsection{Reconstruction Efficiency}
Our method demonstrates significant improvements in computational efficiency compared to traditional sorting-based approaches. Once trained (requiring approximately 15 minutes), our network generates full 3D volumes for any respiratory state in real-time ($<$1 second per volume). This represents a dramatic improvement over conventional methods that require approximately 5 hours per case for non-rigid registration-based reconstruction~\cite{peteani2024retrospective}, making real-time clinical applications feasible.

\begin{figure*}
  \centering
  \includegraphics[width=0.9\textwidth]{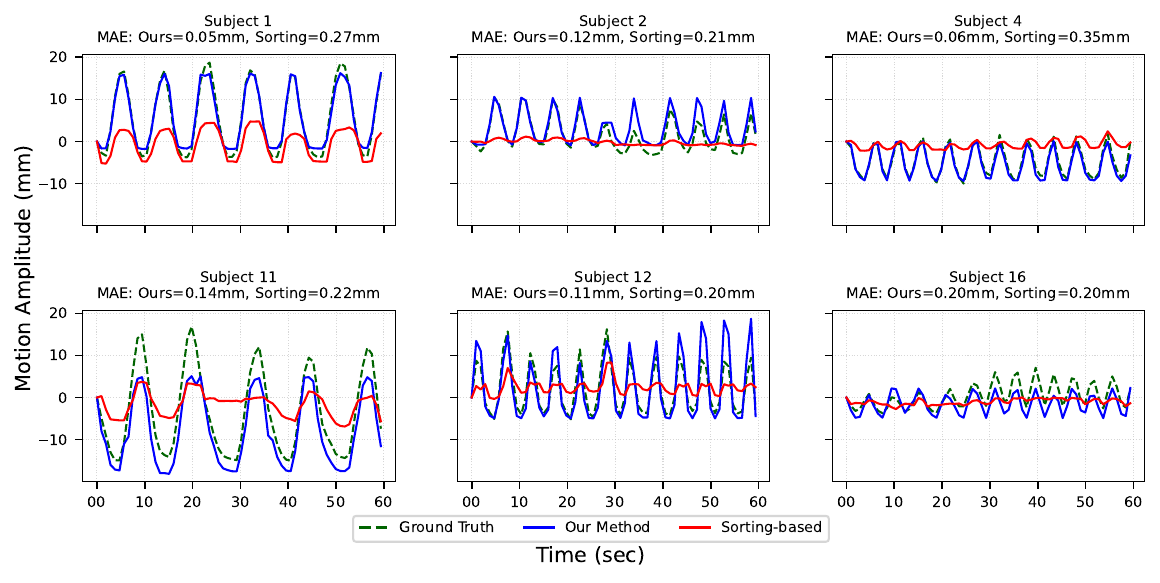}
  \caption{Motion trajectories extracted from 5 diaphragm landmarks using co-tracker from (a) ground truth navigator slices, (b) our INR-reconstructed navigators, and (c) traditional sorting-based reconstruction for six representative volunteers. Peak values generally correspond to inhalation phases, while valley values represent exhalation phases. }
  \label{fig:motion_track}
\end{figure*}

\begin{figure*}
    \centering
    \includegraphics[width=0.85\textwidth]{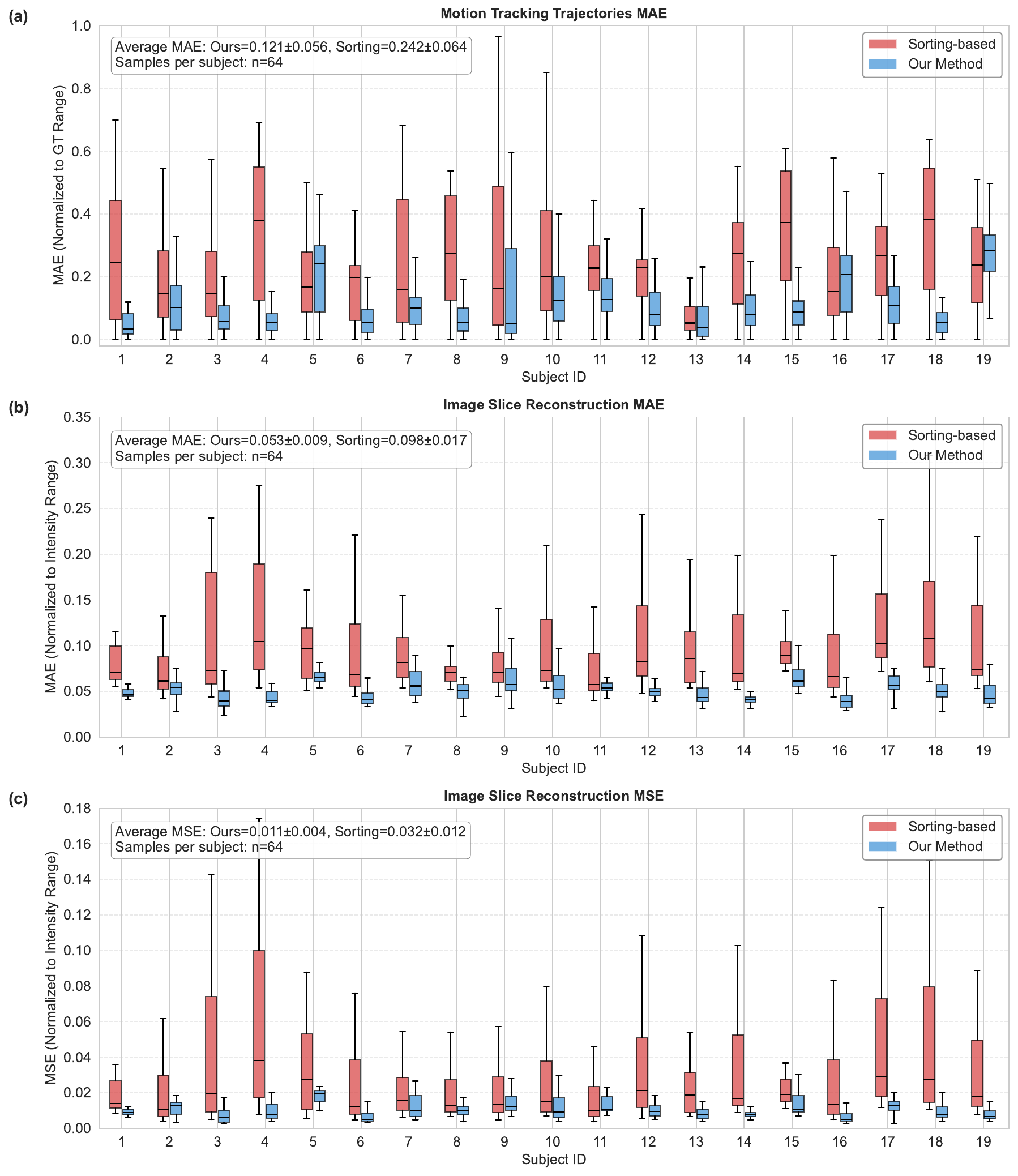}
    \caption{Comparison across 19 test subjects (64 slices each). (a) Motion tracking MAE; (b) Image MAE; (c) Image MSE. Our method (blue) outperforms the sorting-based baseline (red) in all metrics.}
    \label{fig:all_cases_boxplots}
\end{figure*}

\begin{figure*}[t]
  \centering
  \begin{minipage}[t]{0.48\textwidth}
    \centering
    \includegraphics[width=\linewidth]{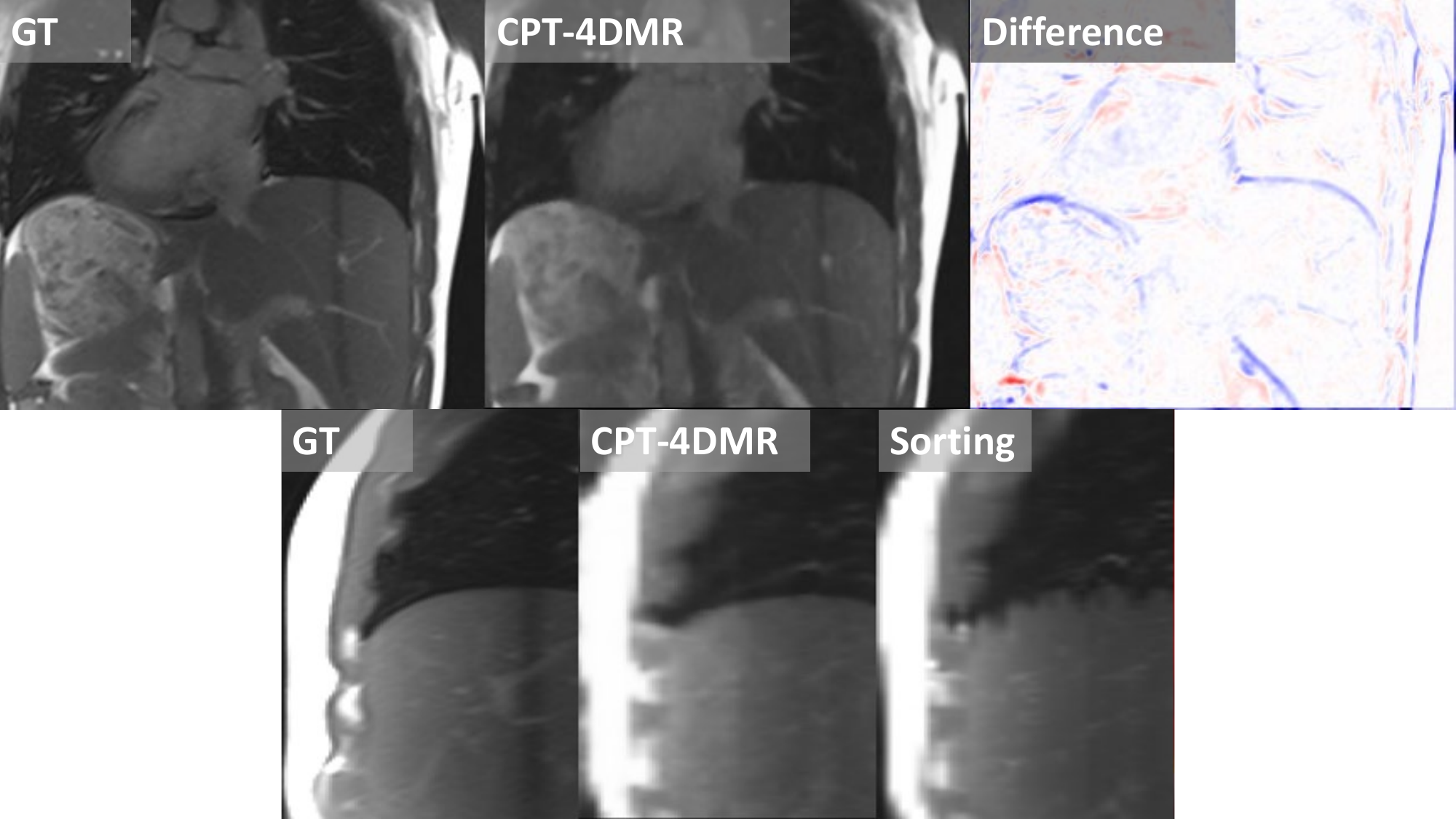}
    \vspace{2pt}
    \small Good case scenario 1 (v3)
    \vspace{6pt}

    \includegraphics[width=\linewidth]{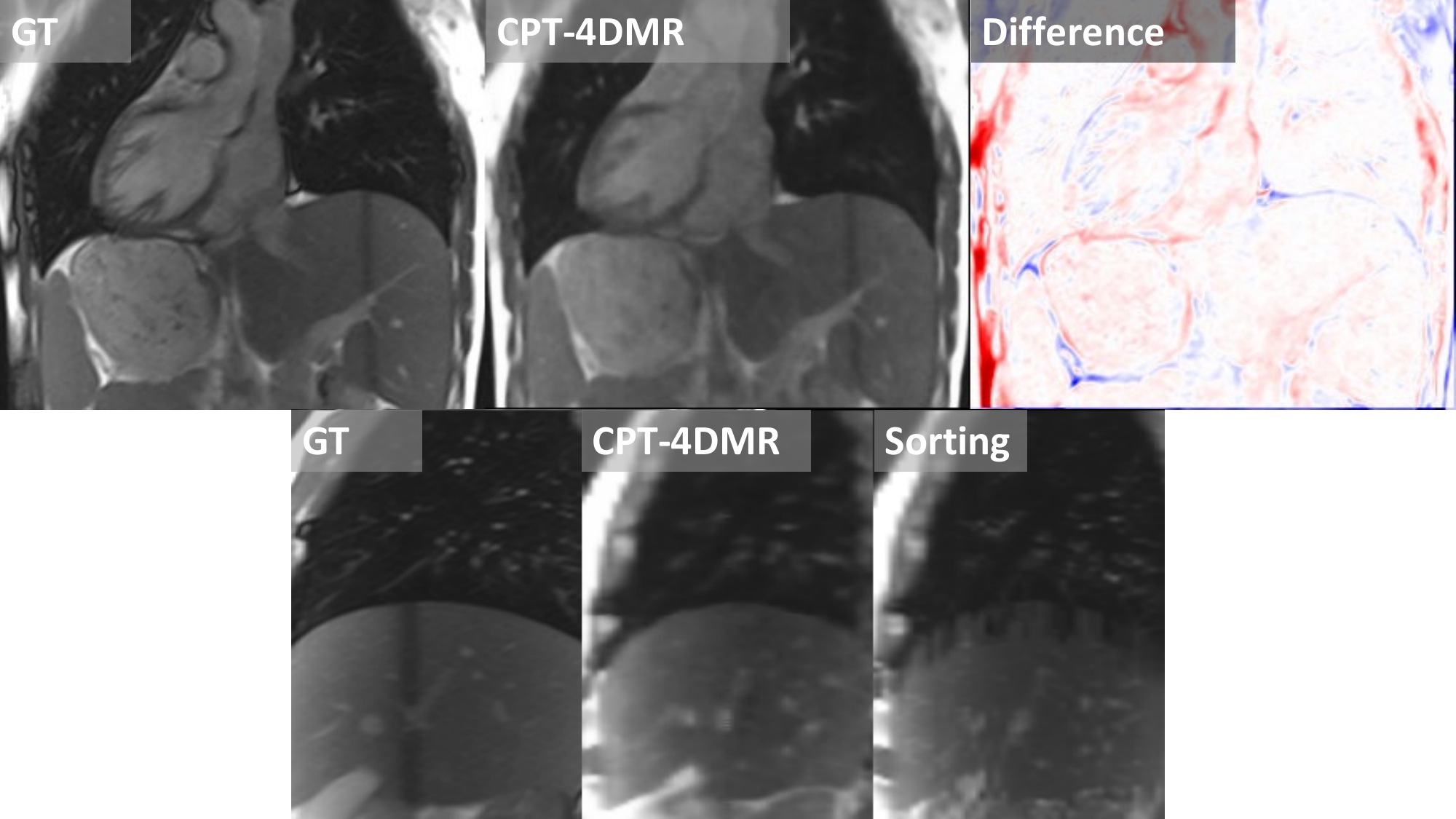}
    \vspace{2pt}
    \small Good case scenario 2 (v12)
  \end{minipage}
  \hfill
  \begin{minipage}[t]{0.48\textwidth}
    \centering
    \includegraphics[width=\linewidth]{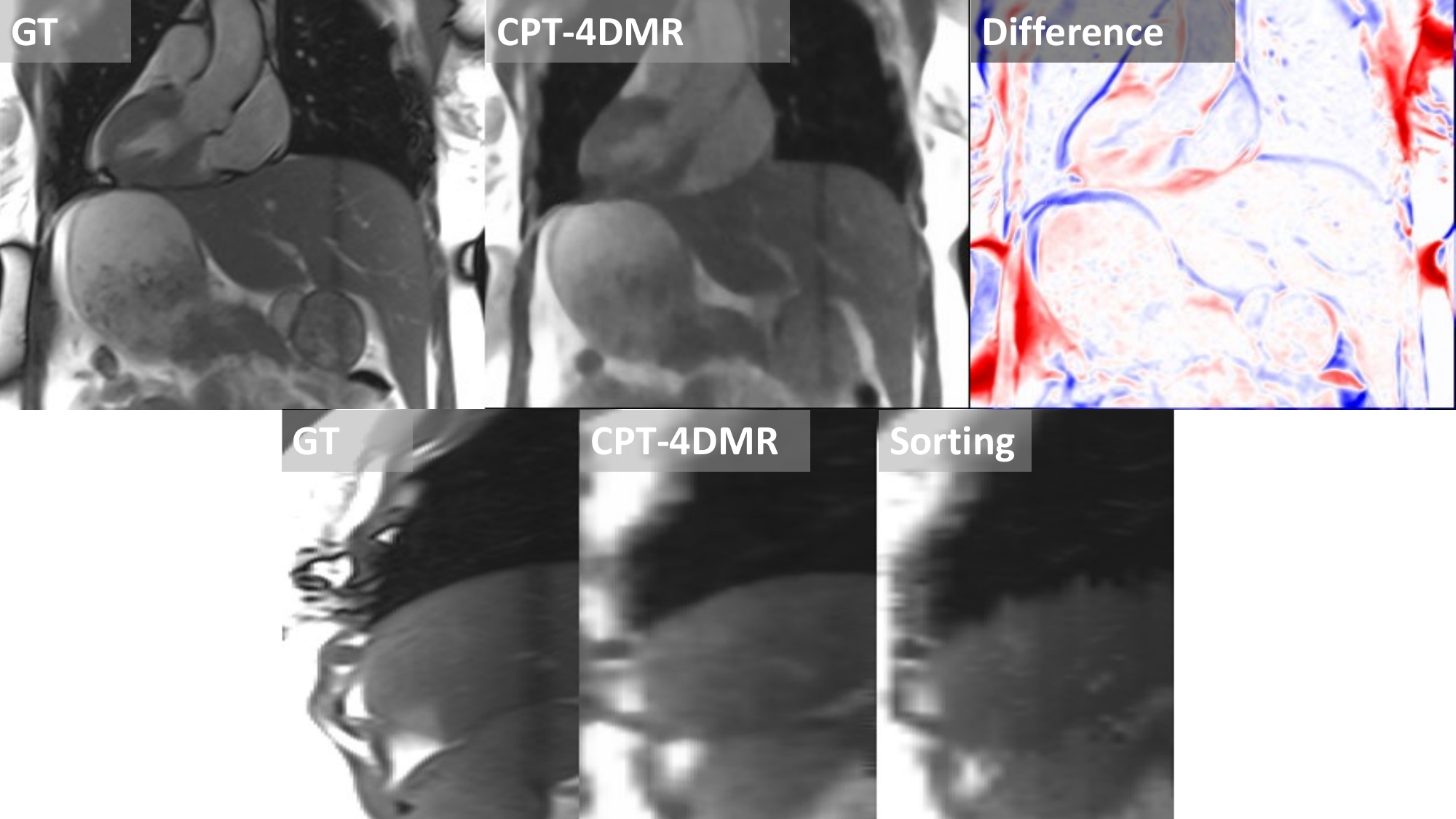}
    \vspace{2pt}
    \small Bad case scenario 1 (v5)
    \vspace{6pt}

    \includegraphics[width=\linewidth]{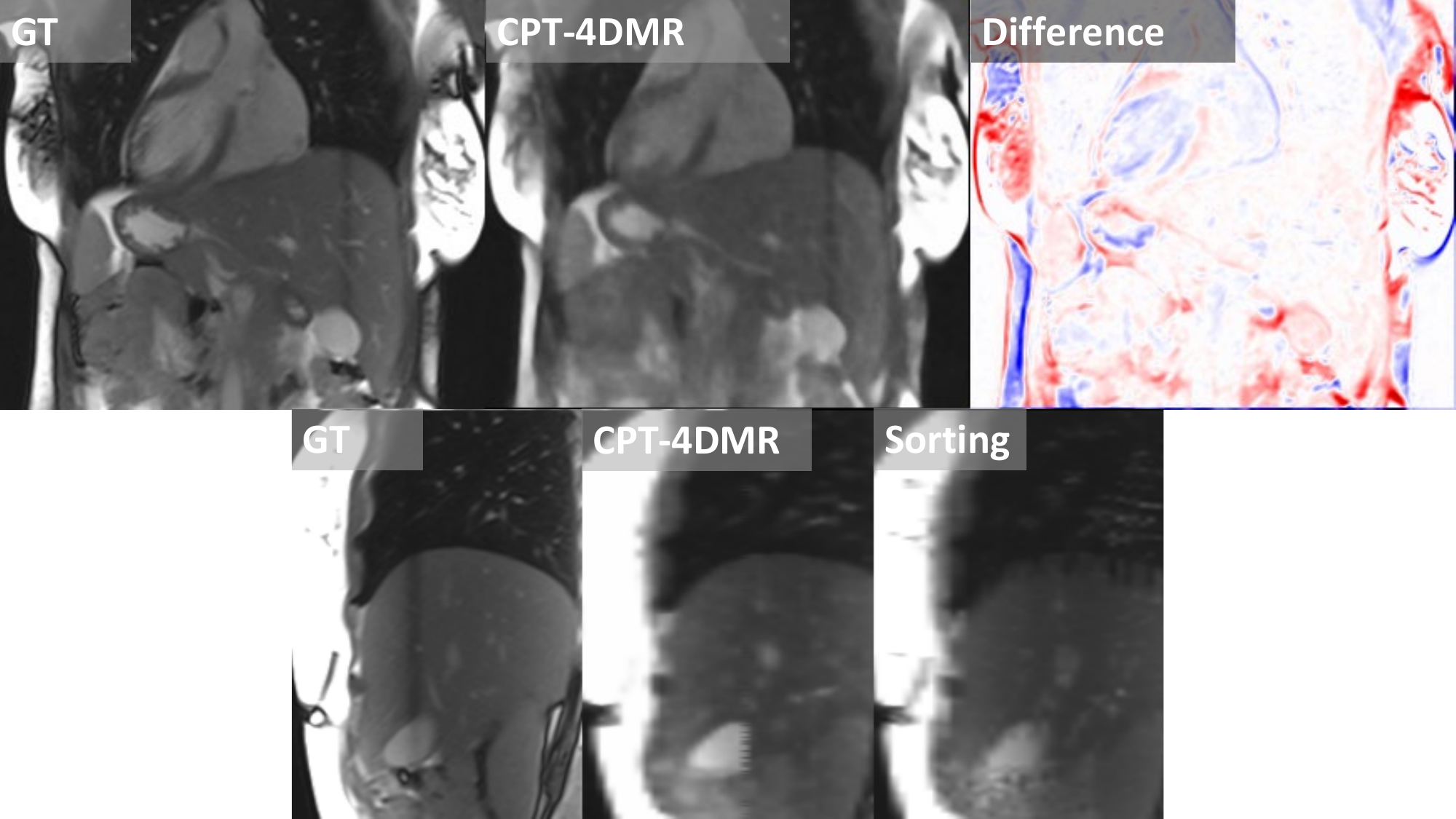}
    \vspace{2pt}
    \small Bad case scenario 2 (v9)
  \end{minipage}

  \caption{Comparison of reconstructed 3D volumes in image slices (first row of each sub-figure) and navigator slices (second row of each sub-figure), showing two relatively good cases (left) and two relatively bad cases (right). Each panel contains ground truth, INR reconstruction, and difference or sorting results.}
  \label{fig:reconstruction_part1}
\end{figure*}


\begin{figure}[t!]
   \centering
   \includegraphics[width=\columnwidth]{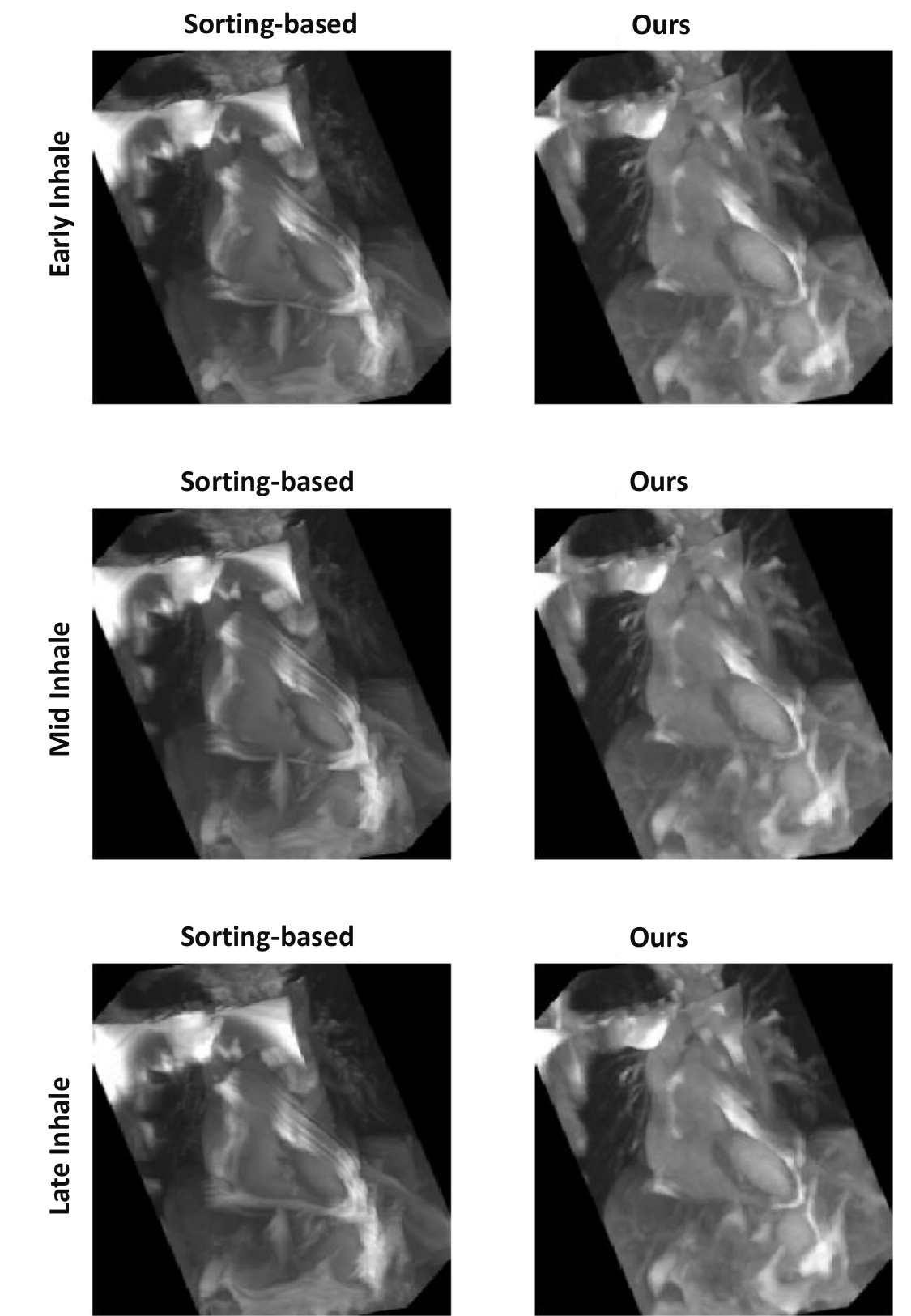}
   \caption{Maximum intensity projections (MIP) of the reconstructed volumes, compare with baseline on the left side.}
   \label{fig:vessel_continuity}
\end{figure}

\begin{figure*}
   \centering
   \includegraphics[width=1.0\textwidth]{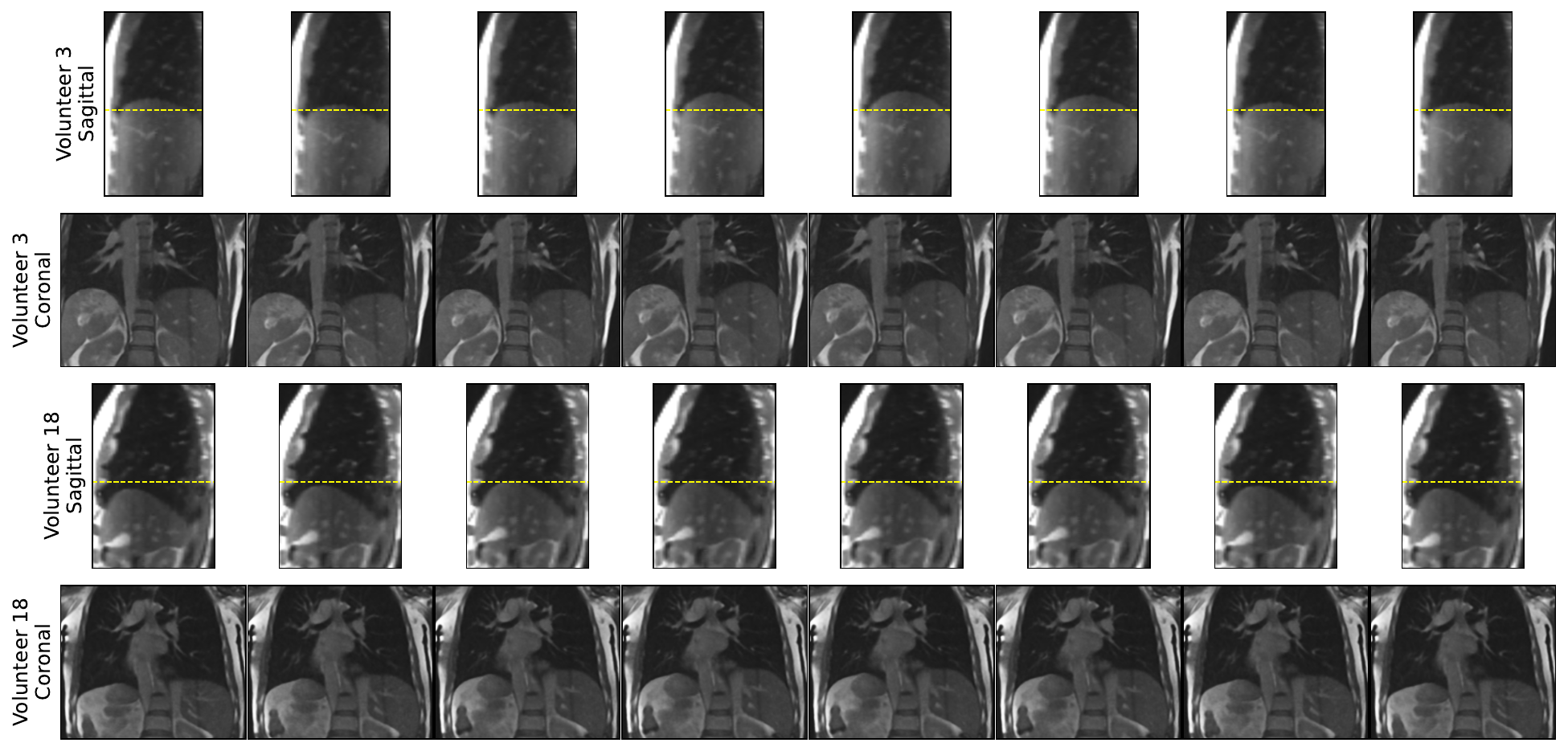}
   \caption{A video of a reconstructed breathing cycle for volunteer 3 and 18. Temporal evolution of reconstructed volumes during a breathing cycle, highlighting the continuity and physical plausibility of the deformations.}
   \label{fig:temporal_sequence}
\end{figure*}

\begin{figure}[t!]
   \centering
   \includegraphics[width=\columnwidth]{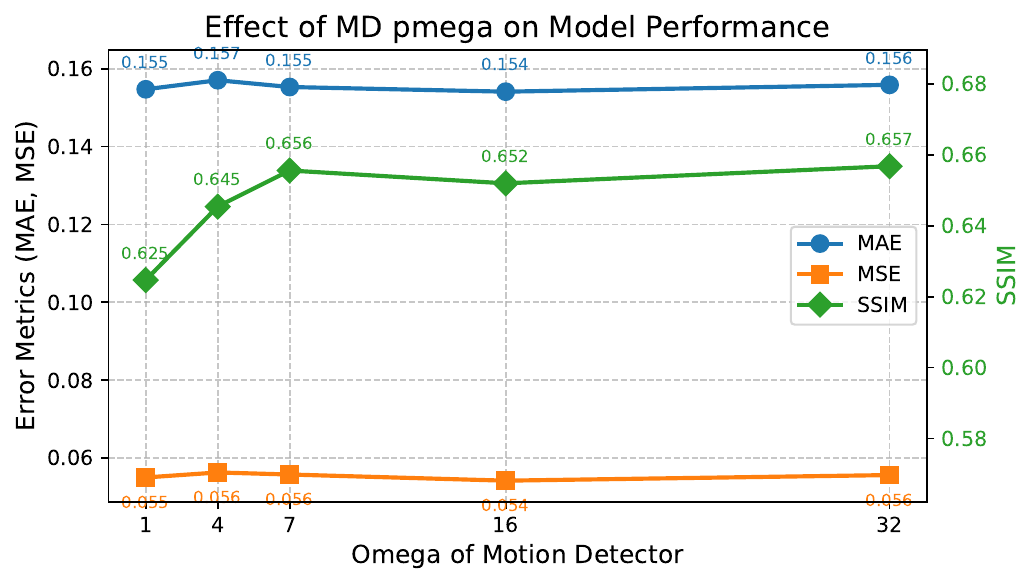}
   \includegraphics[width=\columnwidth]{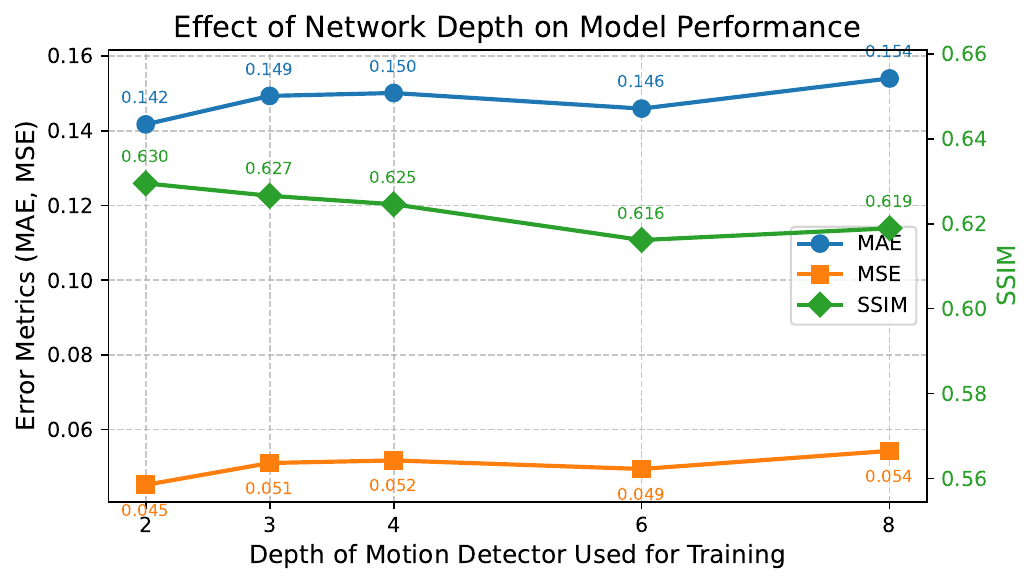}
   \includegraphics[width=\columnwidth]{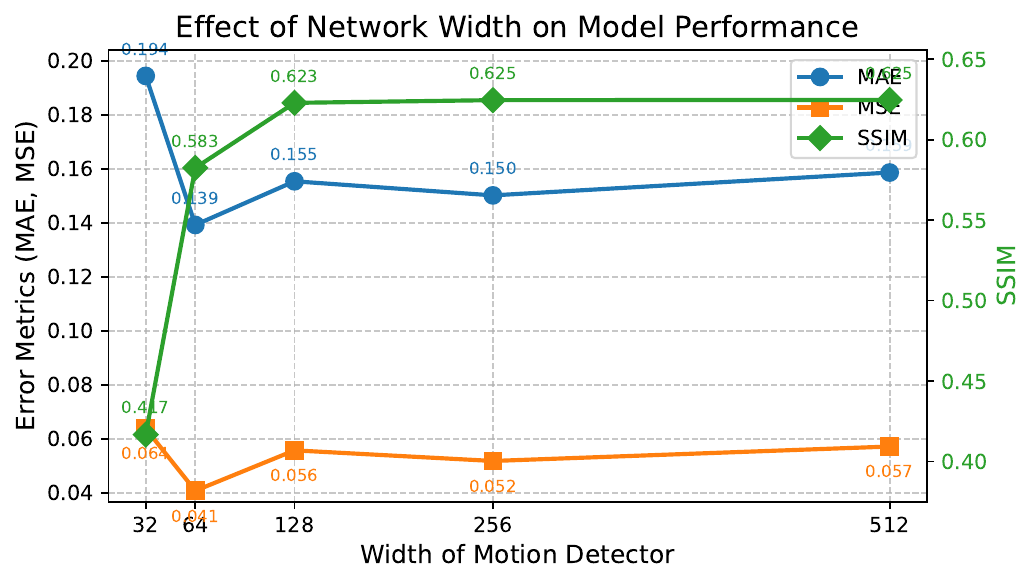}
   \includegraphics[width=\columnwidth]{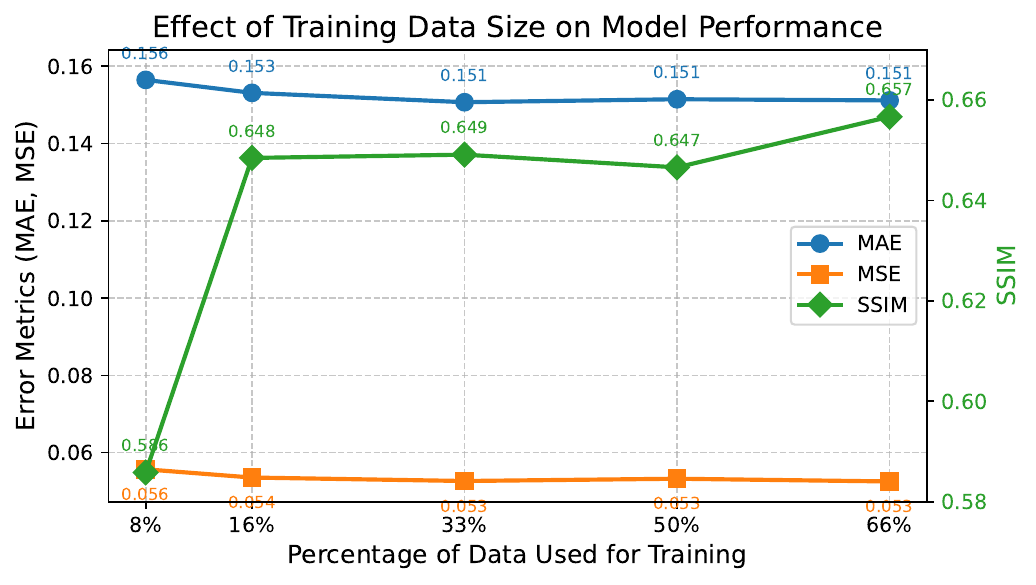}
   \caption{Ablation studies over different designs of network.}
   \label{fig:training_length}
\end{figure}

\subsection{Respiratory Motion Accuracy}

\textbf{Motion Trajectory Analysis}: We evaluated motion accuracy by comparing reconstructed navigator slices with ground truth images throughout complete breathing cycles. Figure~\ref{fig:motion_track} shows the motion trajectories extracted from ground truth navigator slices, our INR-reconstructed navigators, and traditional sorting-based reconstruction for three representative volunteers. Our method demonstrates superior tracking of respiratory motion patterns, closely following ground truth trajectories while sorting-based approaches show significant inconsistencies.


\textbf{Quantitative Motion Assessment}: 
Figure \ref{fig:all_cases_boxplots}(a) presents comprehensive boxplots comparing our method against sorting-based reconstruction across all 19 test cases. The analysis includes: (1) Motion tracking trajectory MAE (top panel), (2) Image slice reconstruction MAE (middle panel), and (3) Image slice reconstruction MSE (bottom panel). Our method consistently achieves lower error values across nearly all subjects, with average normalized MAE of 0.121±0.056 compared to 0.242±0.064 for sorting-based methods in motion tracking.

\subsection{Volumetric Reconstruction Quality}
\textbf{Quantitative Performance}:
Table~\ref{tab1} and Figure \ref{fig:all_cases_boxplots}(b)(c) summarize the quantitative evaluation comparing our INR-based method with traditional sorting-based reconstruction. Our method’s errors are about 50\% lower than those of the traditional approach (e.g., image MAE 0.05 vs 0.10), demonstrating a significant improvement in fidelity. While our evaluation shows overall improved image quality, it should be noted that conventional 4D-MRI reconstruction typically has fewer slices available during inhalation phases compared to exhalation. Our INR-based approach may help address this limitation by providing continuous representation, though a detailed phase-specific quality analysis would require additional temporal binning studies.

\textbf{Visual Quality Assessment}:
Figure \ref{fig:reconstruction_part1} presents detailed comparisons of reconstructed 3D volumes showing (a) ground truth, (b) our method, (c) difference maps, alongside comparisons with sorting-based reconstruction. Our method generates continuous and detailed volumes that closely match ground truth with minimal reconstruction artifacts.

\textbf{Maximum Intensity Projections}: 
Figures \ref{fig:vessel_continuity} demonstrates the continuity of vessel structures through maximum intensity projections (MIP) of reconstructed volumes for volunteers v2, v15, and v18. Unlike traditional methods that rely on discrete slices, our approach generates inherently continuous volumes with consistent anatomical structures across all spatial dimensions.

\textbf{Temporal Consistency}: Figure \ref{fig:temporal_sequence} shows a complete breathing cycle reconstruction for two volunteers, demonstrating the temporal evolution of reconstructed volumes with maintained anatomical plausibility and smooth deformation patterns throughout the respiratory cycle.

\subsection{Ablation Studies}
Figure ~\ref{fig:training_length} presents comprehensive ablation studies examining the effects of: (1) Motion Representation Network omega parameter, (2) Network depth, (3) Network width, and (4) Training data size. Key findings include:
\begin{itemize}
    \item \textbf{Omega parameter:} Optimal performance is achieved with omega values between 1--7, while higher values lead to increased noise sensitivity.
    \item \textbf{Network depth:} 3--4 layers provide the optimal balance between model capacity and overfitting prevention.
    \item \textbf{Network width:} 128--256 units yield the best performance, with diminishing returns beyond 256 units.
    \item \textbf{Training data size:} Performance stabilizes after using 33\% of the available data, suggesting efficient data utilization.
\end{itemize}

\textbf{Template Learning}: Figure \ref{fig:training_progress} demonstrates the progressive improvement of template quality during training, showing how anatomical details become increasingly refined over training epochs.

\section{Discussion}

This study introduces a template-free neural representation framework that jointly learns anatomy and temporal deformations for four-dimensional MRI reconstruction. By eliminating phase binning and pre-acquired reference volumes, the method reduces computational overhead and achieves sub-second volumetric inference, meeting the stringent latency demands of adaptive radiotherapy workflows.

A major strength of our framework lies in its flexibility with respect to respiratory surrogate signals. Unlike conventional methods that require explicit modeling of the relationship between surrogate signals and anatomical motion, our neural representation approach implicitly learns the mapping between any consistent temporal signal and respiratory states through data-driven optimization. Although navigator slices were employed in our experiments, the Temporal Motion Network can adapt to other surrogates, such as external abdominal markers or data-driven self-gated signals \cite{han2018respiratory, vazquez2019automatic}, by automatically discovering the underlying relationship during training. This implicit learning capability means that even surrogate signals with complex or non-obvious relationships to respiratory motion can effectively drive accurate reconstruction, as the network learns to decode respiratory state information from any temporally consistent input signal. This adaptability broadens clinical applicability, particularly in MR-guided radiotherapy setups where external devices may be impractical or introduce additional complexity. One limitation of our approach is the assumption of smooth, continuous motion fields. In reality, lung motion includes sliding boundaries along the chest wall, which may not be perfectly captured by our smooth deformation model. While we apply Jacobian determinant regularization to maintain physical plausibility and prevent tissue folding, incorporating sliding boundary constraints (such as Total Variation regularization) could potentially improve motion field accuracy near organ boundaries in future work.

Our method also addresses key limitations of conventional 4D-MRI workflows. Traditional phase-sorting approaches assume periodic breathing and reconstruct only a single averaged cycle, failing to capture inter-cycle variability and introducing artifacts when breathing is irregular \cite{van2019novel}. Template-based methods further rely on fixed anatomical references, which may not reflect patient-specific motion and often necessitate extra scans or gating procedures. These factors hinder accurate motion modeling and prolong reconstruction times (up to 5 hours in some workflows \cite{peteani2024retrospective}) rendering them unsuitable for real-time use.

In contrast, our implicit neural representation (INR)-based model parameterizes anatomy and motion as continuous functions, enabling flexible sampling of arbitrary respiratory states. INR has recently emerged as a promising paradigm for medical image reconstruction and registration, offering resolution-independent, memory-efficient modeling of dynamic anatomy \cite{molaei2023implicit}. Several studies have demonstrated the superior ability of INRs to represent spatio-temporal deformations and sparse-view reconstructions while preserving anatomical fidelity \cite{feng2025spatiotemporal, byra2023exploring}.

Beyond reconstruction performance, the clinical implications of real-time volumetric inference are considerable. Our framework demonstrates that diaphragm motion, extracted from navigator slices and used as a respiratory surrogate signal, can effectively drive accurate reconstruction of entire thoracic and abdominal volumes throughout the breathing cycle. Real-time 4D imaging based on such diaphragm-driven motion surrogates enables motion-adaptive strategies in radiotherapy, allowing tighter margins and reducing dose to healthy tissue. Multi-institutional studies have confirmed that real-time adaptive radiotherapy significantly reduces dosimetric error compared to static plans, particularly in highly mobile targets such as the lung and liver \cite{colvill2016dosimetric}. 
MRI-guided treatment systems such as MR-Linac further underscore the need for rapid imaging solutions that can continuously monitor and adapt to intra-fractional motion \cite{beaton2019rapid}. Nevertheless, these platforms face technical challenges including magnetic field distortion, processing delays, and workflow complexity, which demand highly efficient reconstruction techniques \cite{thorwarth2021technical}. One current limitation of our framework is the use of 7 mm slice thickness, which restricts the spatial resolution of thin anatomical structures such as small vessels or bronchi. While our method compensates for sparse inter-plane sampling via continuous modeling, thinner slices or isotropic acquisitions would further enhance anatomical fidelity and motion quantification accuracy.

Several enhancements could further strengthen our framework. Incorporating a brief 3D breath-hold scan at the beginning of each session may refine the learned template while preserving the advantages of continuous modeling. Additionally, self-gating strategies that derive surrogate signals directly from image data, such as those based on k-space center tracking or motion-resolved reconstructions, have shown promise in eliminating external dependencies \cite{han2018respiratory, vazquez2019automatic}. Finally, future developments could enable online learning during acquisition, where the neural networks continuously update their parameters as new slices are acquired, allowing the anatomical template and motion model to progressively refine throughout the imaging session. Recent studies have demonstrated that such iterative refinement approaches can significantly improve reconstruction speed and image fidelity \cite{celicanin2015simultaneous, gulamhussene2023transfer}.





\section{Conclusion}
We have introduced a template-free, real-time 4D-MRI reconstruction framework based on implicit neural representations that learns its own anatomical reference volume during training, obviating extra static templates or synthetic data. Joint optimization of this internal template and the deformation fields yields accurate reconstruction across varied breathing patterns, and once trained the system delivers volumetric reconstructions in under one second, which is fast enough for real-time adaptive radiotherapy workflows. Future work will focus on eliminating the surrogate-signal requirement and enabling slice-by-slice real-time updates, further broadening clinical applicability and advancing high-precision temporal reconstruction in radiotherapy.

\section*{Acknowledgement}
This research was supported by the project "Increased Precision for Personalized Cancer Treatment Delivery Utilizing 4D Adapted Proton Therapy (EPIC-4DAPT)," funded by the Swiss National Science Foundation (SNSF) under grant number 212855, and by the project "Ultrasound guided motion mitigation of proton therapy in the lung," funded by the Swiss National Science Foundation (SNSF) under grant number 163330. Some preliminary data processing tools were developed with support from Personalized Health and Related Technologies (PHRT) as part of the interdisciplinary doctoral grant (iDoc 2021-360) of the ETH Domain, Switzerland. We thank Prof. Oliver Bieri for consulting on 4D MR sequence development. We acknowledge Michael Zorneth and the CPT MTRA group for assisting in acquiring 4D MR data. Orso Pusterla acknowledged funding from the Swiss Cystic Fibrosis Society (CFCH).
\bibliographystyle{elsarticle-num} 
\bibliography{mybibliography}
 

\end{document}